\pdfoutput=1
\documentclass[11pt]{article}
\usepackage{xcolor}
\usepackage{booktabs}
\usepackage{multirow}
\usepackage{EMNLP2023}

\usepackage{times}
\usepackage{latexsym}
\usepackage{amsmath} 

\usepackage[T1]{fontenc}
\usepackage[utf8]{inputenc}

\usepackage{microtype}

\usepackage{inconsolata}

\usepackage{graphicx}

\title{{\bf RCEM}: Robust Conversational Search EMbedder in Distributional Shift}

\author{Kilho Son$^{1,}$\thanks{$^{*}$ Corresponding author.} ,  \quad Bo-June Paul Hsu$^{1}$,  \quad  Cha Zhang$^{1}$,  \quad  Dinei Florencio$^{2}$ \\
        $^{1}$\texttt{\{kilhoson, paulhsu, chazhang\}@microsoft.com}, $^{2}$\texttt{dinei@dflorencio.com}\\ \\
        $^{1}$Microsoft Corporation
        }

\begin{document}
\maketitle

\begin{abstract}
We propose RCEM, a Robust Conversational search EMbedder that augments text embedders with LLM query reformulation capabilities while approximately preserving their original generalization performance. Unlike prior conversational dense retrieval approaches that learn direct conversation-to-passage matching, RCEM aligns conversations, prepended by special token, to LLM-rewritten queries, while attempting to preserve the original embedding space. The approximately preserved embedding space automatically maps the rewritten queries to the relevant passages. As a result, RCEM (1) reduces overfitting by aligning conversations to shorter rewritten queries rather than full passages, (2) eliminates the need for conversation-to-passage relevance labels for training, and (3) approximately preserves the original embedding space, enabling conversational retrieval over indexes built by the base embedder without re-indexing. Extensive experiments show that RCEM consistently outperforms prior approaches, improving up to $30\%$ retrieval accuracy under distributional shift.

\end{abstract}

\begin{figure}[ht]
\centering
\includegraphics[trim=0.5cm 5cm 11.5cm 0, width=\columnwidth]{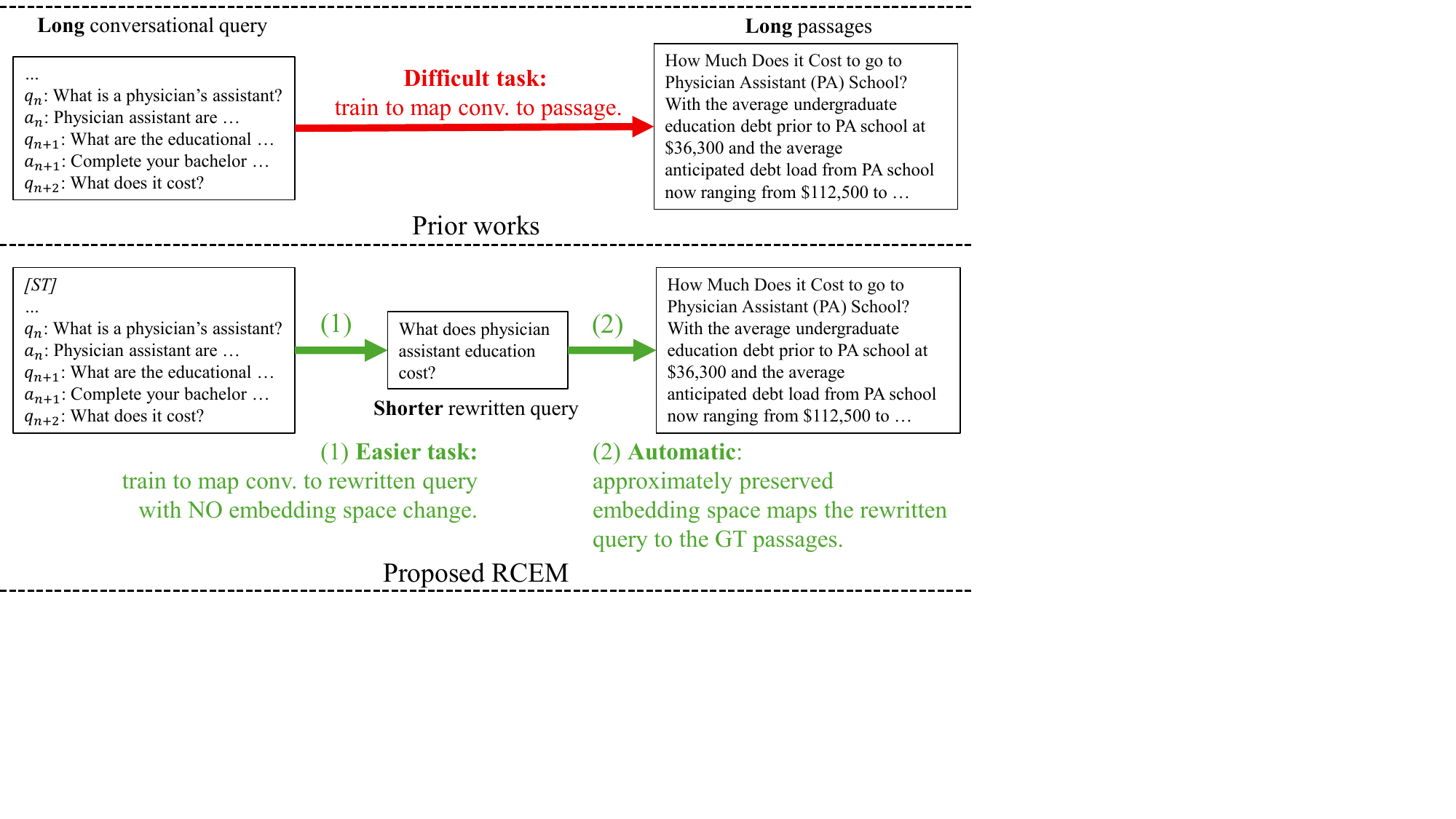}
\caption{{\bf Overview.} 
RCEM is trained to map conversational queries (prepended by a special token, \emph{[ST]}) to LLM-rewritten queries and preserve its original embedding space, which naturally supports retrieval of relevant passages. During inference, \emph{[ST]} triggers RCEM to implicitly convert a conversational query into a clear LLM-rewritten query and computes its embedding in one-step inference, without running an LLM model. Unlike prior methods that directly align long conversations with long passages, the simplified objective reduces overfitting. The approximately preserved embedding space makes RCEM compatible with existing indexes and only activates conversational search when the special token is presented. Otherwise, RCEM behaves similarly to the original embedder. Moreover, RCEM removes the need for expensive conversational query-to-passage relevance annotations for training, enabling greater scalability.
}
\label{fig:overview}
\end{figure}

\section{Introduction}
~\label{sec:intro}
% \footnotetext[1]{Corresponding author}
Conversational search is a core component of retrieval-augmented generation (RAG) systems, where users interact with AI assistants through multi-turn dialogue rather than single queries. In this setting, queries are often context-dependent and omit previously stated information, making user inputs vague, ambiguous, or incomplete~\cite{gao2023neural}. Resolving such contextual dependencies is, therefore, essential for retrieving relevant passages and generating accurate responses.

% Conversational search has become a fundamental component of retrieval augmented generation (RAG) systems, where users interact with AI assistants through multi-turn dialogue rather than isolated queries. In these interactions, user queries are often context-dependent and omit information mentioned in previous turns which makes user inputs often vague, ambiguous or incomplete

A common approach is to first rewrite conversational queries into standalone queries using a large language model (LLM), which can then be embedded and used for dense retrieval. Recent advances in query reformulation have significantly improved retrieval accuracy in multi-turn settings~\cite{mao-etal-2023-llm4cs,mo-etal-2023-convgqr,mo-etal-2024-chiq,adarewriter}. These LLM-based pipelines, however, incur expensive generation cost at inference time, leading to high latency and computational cost, which limits scalability in production. Conversational dense retrieval methods instead aim to generate embeddings directly from conversations, avoiding LLM calls at inference~\cite{emnlp2021, emnlp2025}. These approaches finetune embedders to align conversational queries with relevant passages, requiring high-quality relevance labels. Yet, mapping long conversational queries to long passages is an inherently challenging train objective, making the models prone to overfitting.

% A common solution is to rewrite conversational queries into clear standalone queries using large language models (LLMs) before retrieval. The rewritten query can be converted to embedding that enables conversational dense retrieval. Recent advances in conversational query reformulation have substantially improved retrieval effectiveness in multi-turn settings~\cite{mao-etal-2023-llm4cs,mo-etal-2023-convgqr,mo-etal-2024-chiq,adarewriter}. LLM-based rewriting pipelines, however, require expensive generation during inference, introducing substantial latency and computational cost which hinder cost effective and efficient production. Conversational dense retrieval methods directly generate relevant embedding from the conversation attempting to remove LLM calls in inference time~\cite{emnlp2021, emnlp2025}. They finetune embeder to align the long conversation to long passage, which requires high good quality of relevance label. Mapping long conversation to long passage reduce is relatively difficult task which makes the model fall to overfitting. % robustness under distributional shift.

We propose RCEM, a {\bf R}obust {\bf C}onversational search {\bf EM}bedder that \emph{additionally} learns LLM-based query reformulation capability for conversational search, while preserving the original embedding space and generalization performance. Unlike prior conversational dense retrieval methods~\cite{emnlp2021, emnlp2025}, which directly map conversational queries to ground-truth passages, RCEM aligns conversational queries (prepended by a special token) with LLM-rewritten queries internally and generates their embeddings, while preserving the original embedding space. Because the embedding space is approximately preserved, the resulting embeddings naturally retrieve relevant passages. By aligning conversational queries to shorter LLM-rewritten queries rather than long passages, RCEM simplifies the learning task, reducing overfitting. Maintaining the original embedding space enables conversational retrieval on existing indexes built by original embedder, removing the need for index rebuilding. And conversational search is only triggered for inputs with the special token. Otherwise, the model behaves similarly to the original embedder which increases compatibility with existing search system. Furthermore, RCEM eliminates the expensive cost for high-quality conversational query-to-passage relevance annotations for training, improving scalability (See Figure~\ref{fig:overview}). Extensive experiments show that RCEM consistently outperforms prior work, with particularly large gains under distributional shift.

\section{Related Works}
~\label{sec:related_works}
Existing conversational search approaches generally fall into two categories: conversational query reformulation (CQR) and conversational dense retrieval (CDR). CQR rewrites context-dependent conversational queries into standalone queries prior to sparse or dense retrieval, while CDR directly encodes conversational context into dense embeddings for retrieval.

\paragraph{Conversational Query Reformulation (CQR).} Recent CQR research has focused on improving query reformulation quality through query expansion, intent preservation, ambiguity resolution, and retrieval-aware optimization.~\citet{mo-etal-2023-convgqr} integrate generative query rewriting with potential answer generation for query expansion, while~\citet{mao-etal-2023-search} introduce search-oriented conversational query editing to better preserve retrieval intent during reformulation.~\citet{mao-etal-2023-llm4cs} further improve retrieval effectiveness by aggregating multiple query rewrites and their hypothetical responses into a single retrieval representation. Subsequent work enhances reformulation through more sophisticated generation pipelines, such as rewrite-then-edit strategies~\cite{ye-etal-2023-enhancing}, ambiguity reduction in conversations prior to rewriting~\cite{mo-etal-2024-chiq}, retrieval-aware training objectives that align reformulation with both sparse and dense retrieval~\cite{lai-etal-2025-adacqr}, and test-time adaptation methods that generate multiple candidate rewrites and select the most effective one for retrieval~\cite{adarewriter}. CQR approaches are modular and retriever-agnostic, allowing seamless integration with existing sparse and dense retrieval systems. They also naturally benefit from advances in LLMs and prompting techniques. However, they incur substantial cost and latency at inference due to repeated LLM calls, making large-scale deployment challenging.

\paragraph{Conversational Dense Retrieval (CDR).}
CDR methods directly encode conversational context into dense retrieval representations.~\citet{emnlp2021, emnlp2025} introduce contextualized query embeddings that incorporate conversation history for conversational retrieval by training to map conversational queries to the relevant passages. Subsequent research improves retrieval effectiveness through context-denoised representations and pseudo-supervision mined from historical turns~\cite{mo-etal-2024-history}, generalized conversational retrieval models based on LLMs and dual-learning objectives~\cite{mao-etal-2024-chatretriever}, and methods that enhance interpretability by decoding dense conversational embeddings into human-readable queries while preserving retrieval performance~\cite{cheng-etal-2024-interpreting}. These approaches benefit from end-to-end optimization and deep integration of conversational context into retrieval representations.
% ~\citet{emnlp2021, emnlp2025} propose contextualized query embeddings that incorporate conversation history into dense retrieval representations for conversational search.~\citet{mo-etal-2024-history} improve conversational search through context-denoised query reformulation and the automatic mining of pseudo-positive and hard-negative supervision signals from historical turns.~\citet{mao-etal-2024-chatretriever} presents ChatRetriever, which adapts LLM for generalized conversational dense retrieval using a dual-learning approach to robustly understand and represent complex, multi-turn conversational contexts.~\citet{cheng-etal-2024-interpreting} improves the interpretability of conversational dense retrieval models by using a customized Vec2Text framework enhanced with external query rewrites to decode opaque session embeddings into human-readable text while preserving original retrieval performance. These methods benefit from end-to-end optimization and deep integration of conversational context within retrieval representations. 
However, existing CDR methods typically rely on contrastive training between conversational queries and relevant passages using high-quality conversation-to-passage relevance annotations. Acquiring such supervision is expensive and challenging, limiting scalability. Furthermore, training on relatively small conversational retrieval datasets can lead to overfitting, while contrastive objectives may alter the original embedding space and reduce compatibility with existing retrieval systems. In contrast, RCEM eliminates the need for conversation-to-passage relevance supervision and explicitly attempts to preserve the original embedding space, improving generalization, scalability, and compatibility with existing retrieval infrastructures.
\section{Proposed RCEM}
~\label{sec:rcem}
\begin{figure*}[t]
\centering
\includegraphics[trim=3cm 7cm 4cm 0, width=1.3\columnwidth]{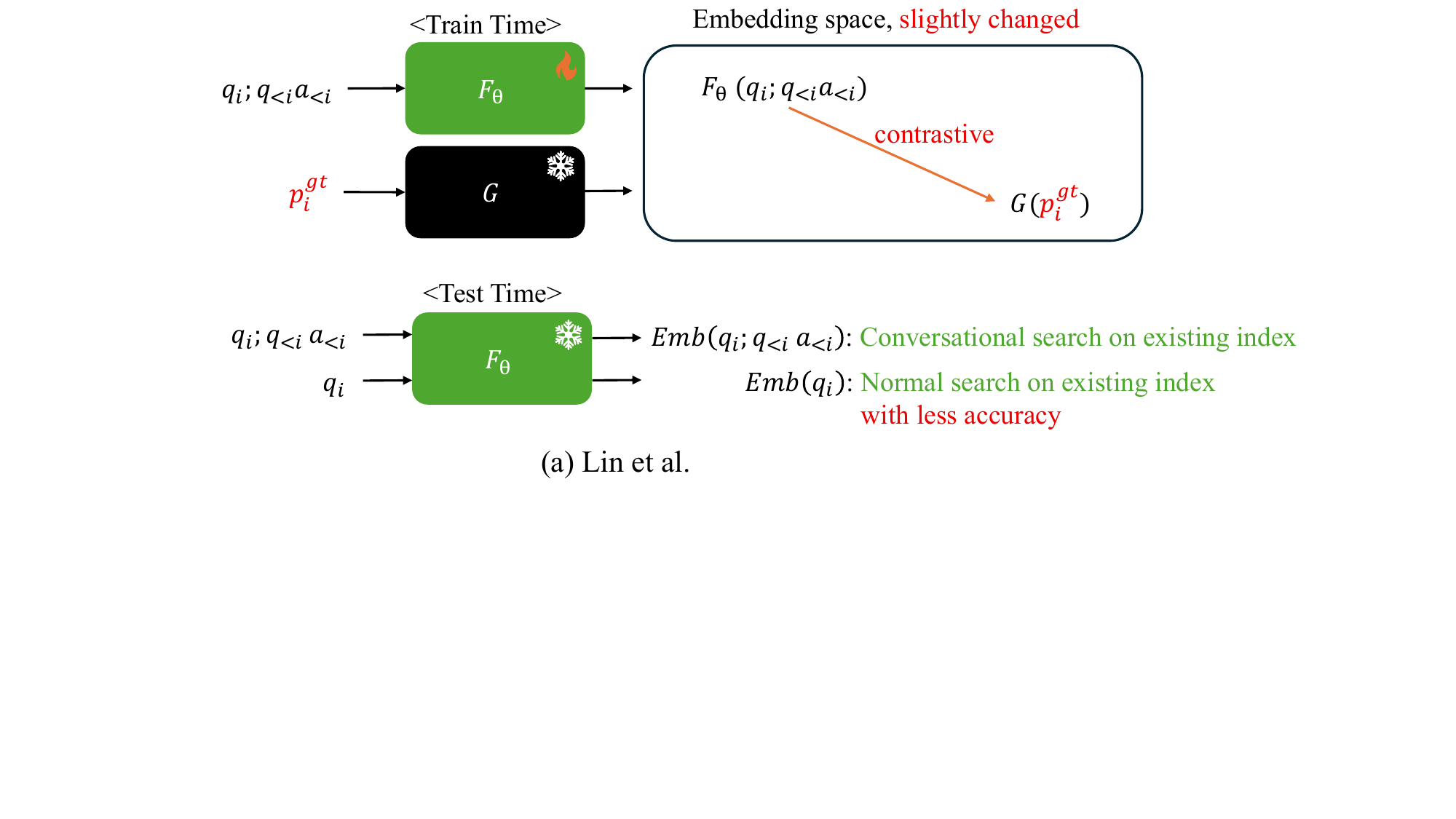}
\includegraphics[trim=3cm 3cm 4cm 0, width=1.3\columnwidth]{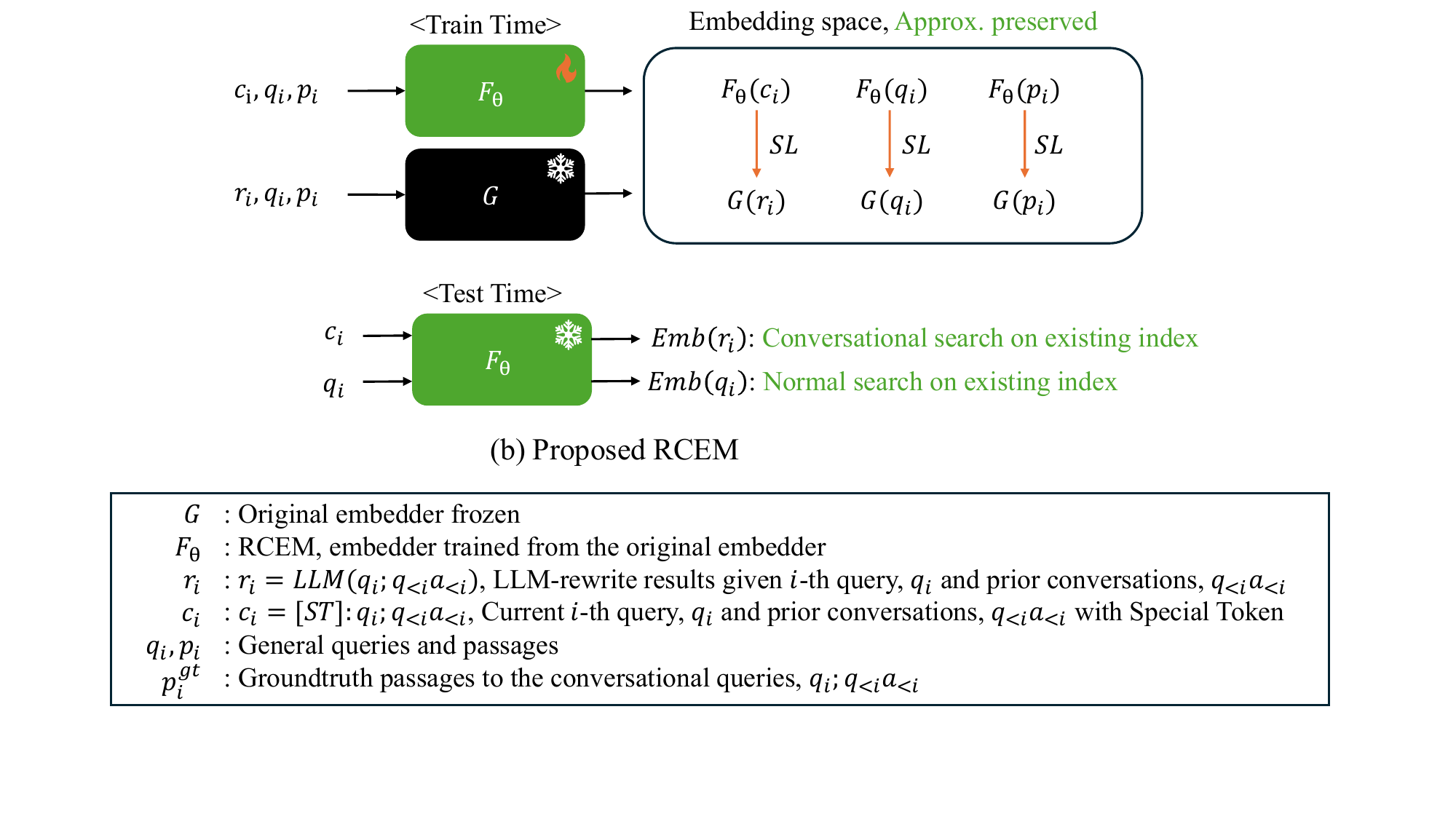}
\caption{{\bf RCEM overview and comparison with prior work.} RCEM first precomputes embeddings of the LLM-rewritten query, $r_i = LLM(q_i; q_{<i}a_{<i})$, as well as general queries $q_i$ and passages $p_i$ using the original frozen embedder $G$. RCEM $F_\theta$ is then trained to map conversational queries prepended by a special token, $F_\theta(c_i)$, to the embedding of the rewritten query produced by the frozen embedder $G(r_i)$ using the proposed Structure Learning (SL) loss. During training, RCEM also aligns general queries and passages with their precomputed embeddings, i.e., $F_\theta(q_i) \rightarrow G(q_i)$ and $F_\theta(p_i) \rightarrow G(p_i)$, to preserve the original embedding space. This design enables a single model to support both conversational search, triggered by the special token, and standard search over the index built by the original embedder. Attempting to preserve the embedding space also serves as a regularizer, making the model more robust to distributional shift. In contrast, ~\citet{emnlp2021} trains the model to map the conversational query embedding $F_\theta(q_i; q_{<i}a_{<i})$ directly to the ground-truth passage $G(p_i^{gt})$, which is a more complicated training task, resulting in overfitting. Unlike~\citet{emnlp2021, emnlp2025}, RCEM does not require conversational-to-passage relevance annotations for training, which are often difficult to obtain accurately. This makes RCEM more scalable, as it reduces the need for expensive and high-quality relevance annotations.}

% RCEM first precomputes embeddings of rewritten query using LLM where input is query and conversation ($q_i;q_{<i}a_{<i}$), $R_i$, general query $q$ and documents $D$ using the original embedder $G$. The same base model, $F$ is trained to maps with Special Token $[ST,q_i;q_{<i}a_{<i}]$ to embedding of $R_i$ with structure learning loss. While training this, RCEM also maps queries and documents to the precomputed embedding, $F(q) \rightarrow G(q), F(D) \rightarrow F(D)$ to maintain the embedding space unchanged. This enables to use the same model for conversational search which is triggered by special token and general search query search on the existing index. This also roles as regularization which enables the model robust to the distributional shift. Whereas, ~\cite{emnlp2021} trains the model to map conversational query, $q_i;q_{<i}a_{<i}$ to groundtruth document $D_{i}^{gt}$ which changes the embedding space. Different from prior works, RCEM does not require conversational query and groundtruth passage mapping for training which often challenging to acquire high quality label which enables to train scalable training due to convenience of data label acquire.}
\label{fig:rcem}
\end{figure*}

We propose RCEM, a robust conversational search embedder that directly outputs an embedding for conversational search. Unlike conventional conversational search pipelines that rely on explicit LLM-based query rewriting during inference, RCEM intrinsically converts a conversational query into an LLM-rewritten query and generates an embedding of the rewritten query in a single step, which reduces inference computational cost and latency for conversational search. RCEM enables this by \emph{additionally} injecting LLM query reformulation capability into the embedder while attempting to preserve the original embedder's capability. RCEM is described in detail in this section.

\paragraph{Setting.}
Let $q_i$ denote the current query at dialogue turn $i$, and let $q_{<i}, a_{<i}$ denote the conversational history. Thus, the conversational query is $q_i;q_{<i}, a_{<i}$. 
% $p_i^{gt}$ is a groundtruth passage of the . $p_i$ and $q_i$ represent a general passage and standalone query respectively.
Existing conversational search systems commonly use an LLM-based rewriter
\begin{equation}
    r_i = LLM(q_i; q_{<i}, a_{<i}),
    \label{equ:1}
\end{equation}
where $r_i$ is an LLM-rewritten query suitable for retrieval.
To distinguish conversational search from standard retrieval, we prepend a special token $[ST]$ to the conversational query,
\begin{equation}
    c_i =
    [ST]: q_i; q_{<i} a_{<i}.
    \label{equ:2}
\end{equation}

% $Q$, $R$, $C$ and $P$ are sets of query ($q_i$), LLM-rewritten query ($r_i$), conversational query prepended with special token ($c_i$) and passages $p$ in the training set, respectively. We define a set of pairs for training RCEM,

% where, $\Omega_Q = $,
% $\Omega_P = $ and $\Omega_R = \{(c_i, r_i)\}_{i=0}^n$

\paragraph{Embedder architecture.}
Two embedders are defined: RCEM $F_\theta(x)$ and the original embedder $G(x)$, where
\begin{equation}
    F_\theta(x) = \mathrm{MLP}_{\mathrm{SELU}}(\mathrm{LoRA}(G(x))).
    \label{equ:3}
\end{equation}

$F_\theta(x)$ is built on top of the frozen $G(x)$ with LoRA adapters and a lightweight two-layer perceptron head with SELU activation~\cite{selu}. The backbone parameters are initialized from $G(x)$, while only the LoRA adapters and the MLP head, parameterized with $\theta$, are updated during training.

\paragraph{Structure Learning (SL).}  $F_\theta(x)$ is trained by minimizing the loss below,
\begin{equation}
    \mathcal{L}(\theta)
    =
    \mathcal{L}_{\mathrm{point}}(\theta)
    +
    \lambda
    \mathcal{L}_{\mathrm{pair}}(\theta),
    \label{equ:4}
\end{equation}

where
\begin{equation}
    \mathcal{L}_{\mathrm{point}}(\theta)
    =
    \sum_{j=0}^{l+m+n}
    \left\|
    F_{\theta}(x_{j,1}) - G(x_{j,2})
    \right\|,
    \label{equ:5}
\end{equation}

% \begin{multline} 
%     \mathcal{L}_{\mathrm{pair}}(\theta) = \sum_{j=0}^{l+m+n} \sum_{k > j}^{l+m+n} \Big\| \left\|F_{\theta}(x_{j,1})-F_{\theta}(x_{k,1})\right\| \\
%     - \left\|G(x_{j,2})-G(x_{k,2})\right\| \Big\|.
%     \label{equ:6}
% \end{multline}
\begin{equation}
    \begin{aligned} 
        \mathcal{L}_{\mathrm{pair}}(\theta) = \sum_{j=0}^{l+m+n} \sum_{k > j}^{l+m+n} \Big\| \left\|F_{\theta}(x_{j,1})-F_{\theta}(x_{k,1})\right\| \\
        - \left\|G(x_{j,2})-G(x_{k,2})\right\| \Big\|.
        \label{equ:6}
    \end{aligned}
\end{equation}

$x_{j,1}$ and $x_{j,2}$ are the first and second elements of the $j$-th pair in a set of pairs,

\begin{align}
    \Omega = \{(q_i, q_i)\}_{i=0}^l  \cup  \{(p_i, p_i)\}_{i=0}^m \cup \{(c_i, r_i)\}_{i=0}^n.
\end{align}

$\Omega$ consists of three subsets of matching pairs, $l$ pairs of general queries ($q_i$) to themselves, $m$ pairs of passages ($p_i$) to themselves, and $n$ conversational queries prepended by $[ST]$ to their LLM-rewritten queries. The last set of matching pairs are for training RCEM to align a conversational queries prepended by the special token toward an LLM-rewritten queries, $F_\theta(c_i) \rightarrow G(r_i)$. This enables RCEM to learn the LLM-based rewriting capability and directly encode the conversational query into the embedding of the rewritten query without explicitly generating $r_i$ during inference. The first two sets of pairs, $\{(q_i, q_i)\}_{i=0}^l  \cup  \{(p_i, p_i)\}_{i=0}^m$ are for preserving the embedding space. While distilling LLM's query rewriting capability into the embedder, RCEM also attempts to preserve the embedding space by mapping $F_\theta(q_i) \rightarrow G(q_i)$ and $F_\theta(p_i) \rightarrow G(p_i)$. 

% Different from~\citet{emnlp2025} that uses pointwise distance for training, our loss also preserves the distance between pairs in training and $\lambda$ controls the contribution of the pairwise structure term. We set $\lambda$ as $1.$ unless otherwise stated. Figure~\ref{fig:rcem} illustrate an overview of the RCEM and comparison with ~\citet{emnlp2021} and Appendix~\ref{app:compare} comare RCEM with ~\citet{emnlp2025}.

Unlike~\citet{emnlp2025}, which optimizes only pointwise embedding distances, RCEM also preserves pairwise structure during training. The hyperparameter $\lambda$ controls the weight of the pairwise structure preservation term, and is set to $1.0$ unless otherwise specified. Figure~\ref{fig:rcem} provides an overview of RCEM and contrasts it with~\citet{emnlp2021}. A more comparison between RCEM and~\citet{emnlp2025} is presented in Appendix~\ref{app:compare}.

This design offers several advantages. First, RCEM eliminates expensive LLM-based rewriting during inference, substantially reducing latency and computational cost. Second, RCEM does not require conversational-query-to-passage relevance supervision, making training more scalable. Finally, the approximately preserved original embedding space serves as a regularizer during training that increases robustness in distributional shift and remains compatible with existing indexes built by the original embedder.

% Unlike prior methods~\cite{emnlp2021, emnlp2025}, RCEM does not require ground-truth passage $p_i^{gt}$ mappings for conversational queries $(q_i; q_{<i}a_{<i})$. This removes the dependence on expensive and difficult-to-obtain conversational relevance annotations, making the training pipeline more scalable and improving robustness under distributional shift. 

\section{Experiments}
\label{sec:experiments}
% \begin{figure}[t]
% \centering
% \includegraphics[width=\columnwidth]{figs/in_domain_comparison.png}
% \caption{{\bf In domain performance comparison}: We train model using QReCC train and evaluate the performance with test data in the same domain with the same setups for fair comparison. Our proposed method outperforms~\citet{emnlp2021} by 3.0\% in Recall@10, 2.7\% in MRR, and 2.5\% in NDCG@3. Note that QReCC has an index size of approximately 54M passages, and Qwen3-0.6B is used as the base embedder.}
% \label{fig:indomain}
% \end{figure}

% \begin{table*}[t]
% \centering
% \small
% \begin{tabular}{lcccccc}
% \toprule
% \multirow{2}{*}{ } &
% \multicolumn{3}{c}{QReCC} &
% \multicolumn{3}{c}{TopiOCQA} \\
% \cmidrule(lr){2-4} \cmidrule(lr){5-7}
% & R@10 & MRR & NDCG@3
% & R@10 & MRR & NDCG@3 \\
% \midrule
% Query Rewrite (GPT-4.1) $\rightarrow$ Embedding& 73.3 & 47.3 & 45.4 & 64.6 & 38.1 & 37.6 \\
% \citet{emnlp2021}& 67.5 & 39.3 & 37.4 & 66.3 & 40.3 & 40.1 \\
% \citet{emnlp2025} & - & 36.8 & - & - & 42.2 & - \\
% Proposed RCEM& {\bf 75.8} & {\bf 49.9} & {\bf 48.3} & {\bf 70.4} & {\bf 43.2} & {\bf 42.9} \\
% \bottomrule
% \end{tabular}
% \caption{{\bf In domain performance comparison.} RCEM consistently outperforms all the other methods, achieving approximately ~10\% higher MRR than~\citet{emnlp2021} on QReCC and 1.0\% higher MRR than~\citet{emnlp2025} on TopiOCQA.}
% \label{tab:larger-model-indomain}
% \end{table*}

\begin{table*}[t]
\centering
\small
\begin{tabular}{llcccccc}
\toprule
\multirow{2}{*}{Type} &
\multirow{2}{*}{Method} &
\multicolumn{3}{c}{QReCC} &
\multicolumn{3}{c}{TopiOCQA} \\
\cmidrule(lr){3-5} \cmidrule(lr){6-8}
& & R@10 & MRR & NDCG@3
& R@10 & MRR & NDCG@3 \\
\midrule
CQR & ConvGQR~\cite{mo-etal-2023-convgqr}
    & 63.5 & 42.0 & 39.1 & 41.8 & 25.6 & 24.3 \\
CQR & LLM4CS, $N=16$~\cite{mao-etal-2023-llm4cs}
    & 64.6 & 43.2 & 40.7 & 55.1 & 35.4 & 34.5 \\
CQR & CHIQ-Fusion~\cite{mo-etal-2024-chiq}
    & 70.7 & 47.2 & 44.2 & 61.6 & 38.0 & 37.0 \\
CQR & AdaCQR+Expansion~\cite{lai-etal-2025-adacqr}
    & 67.3 & 45.8 & 42.9 & 58.4 & 38.5 & 37.6 \\
CQR & AdaRewriter, $N=16$~\cite{adarewriter}
    & 69.8 & 47.5 & 44.7 & 61.9 & 40.3 & 39.7 \\
CQR & Query Rewrite (GPT-4.1) $\rightarrow$ Embedding
    & 73.3 & 47.3 & 45.4 & 64.6 & 38.1 & 37.6 \\
\midrule
CDR & \citet{emnlp2021}
    & 67.5 & 39.3 & 37.4 & 66.3 & 40.3 & 40.1 \\
CDR & \citet{emnlp2025}
    & -- & 36.8 & -- & -- & 42.2 & -- \\
CDR & Proposed RCEM
    & \textbf{75.8} & \textbf{49.9} & \textbf{48.3}
    & \textbf{70.4} & \textbf{43.2} & \textbf{42.9} \\
\bottomrule
\end{tabular}
\caption{\textbf{In-domain retrieval accuracy comparison.} RCEM consistently outperforms both the conversational query reformulation (CQR) and conversational dense retrieval (CDR) methods, achieving $\sim 10\%$ and $\sim 13\%$ higher MRR than ~\citet{emnlp2021} and~\citet{emnlp2025} on QReCC respectively.
% CQR and CDR denote conversational query reformulation and conversational dense retrieval, respectively.
}
\label{tab:larger-model-indomain}
\end{table*}

% \begin{figure*}[t]
% \centering
% \includegraphics[width=1.8\columnwidth]{figs/out_of_domain_8b.png}
% \caption{{\bf Distributional shift performance comparison.} Models are trained on one dataset and evaluate them on another. The proposed RCEM method maintains strong performance under distributional shift, whereas the performance of~\citet{emnlp2021} drops noticeably. In particular, when trained on QReCC and tested on TREC CAsT 2019, RCEM achieves approximately ~30\% higher MRR than~\citet{emnlp2021}.}
% \label{fig:outdomain_8b}
% \end{figure*}
\begin{table*}[t]
\centering
\small
\begin{tabular}{lllccc}
\toprule
Method & Test dataset & Train dataset & R@10 & MRR & NDCG@3 \\
\midrule

 Query Rewrite (GPT-4.1) $\rightarrow$ Embedding & \multirow{3}{*}{QReCC} & -- & 73.3 & 47.3 & 45.4 \\
 \citet{emnlp2021} &  & TopiOCQA & 68.0 & 42.2 & 40.5 \\
 Proposed RCEM &  & TopiOCQA
    & \textbf{76.3} & \textbf{51.3} & \textbf{49.8} \\
\midrule

Query Rewrite (GPT-4.1) $\rightarrow$ Embedding& \multirow{3}{*}{TopiOCQA} & -- & 64.6 & 38.1 & 37.6 \\
\citet{emnlp2021}& & QReCC & 44.1 & 24.3 & 23.4 \\
Proposed RCEM & & QReCC
    & \textbf{65.3} & \textbf{38.6} & \textbf{37.8} \\
\midrule

 Query Rewrite (GPT-4.1) $\rightarrow$ Embedding & \multirow{5}{*}{TREC CAsT 2019} & -- & 90.2 & 73.4 & 46.7 \\
 \citet{emnlp2021}& & QReCC & 76.3 & 48.7 & 26.7 \\
 Proposed RCEM& & QReCC
    & 93.1 & \textbf{77.4} & \textbf{51.8} \\
 \citet{emnlp2021}& & TopiOCQA & 89.0 & 63.4 & 38.1 \\
 Proposed RCEM& & TopiOCQA
    & \textbf{93.6} & 76.2 & 50.5 \\
\midrule

 Query Rewrite (GPT-4.1) $\rightarrow$ Embedding & \multirow{5}{*}{TREC CAsT 2020} & -- & 87.0 & 65.9 & 37.8 \\
 \citet{emnlp2021}& & QReCC & 64.9 & 40.1 & 19.7 \\
 Proposed RCEM& & QReCC & 86.5 & 64.3 & 39.3 \\
 \citet{emnlp2021}& & TopiOCQA & 79.8 & 52.1 & 26.4 \\
 Proposed RCEM& & TopiOCQA
    & \textbf{88.0} & \textbf{68.1} & \textbf{39.5} \\
\bottomrule
\end{tabular}
\caption{{\bf Retrieval accuracy comparison in distributional shift.} Models are trained on one dataset and evaluate them on another. The proposed RCEM method maintains strong performance under distributional shift, whereas the performance of~\citet{emnlp2021} drops noticeably. In particular, when trained on QReCC and tested on TREC CAsT 2019, RCEM achieves $\sim 30\%$ higher MRR than~\citet{emnlp2021}.}
\label{tab:outdomain_8b}
\end{table*}

\begin{figure*}[t]
\centering
\includegraphics[width=1.6\columnwidth]{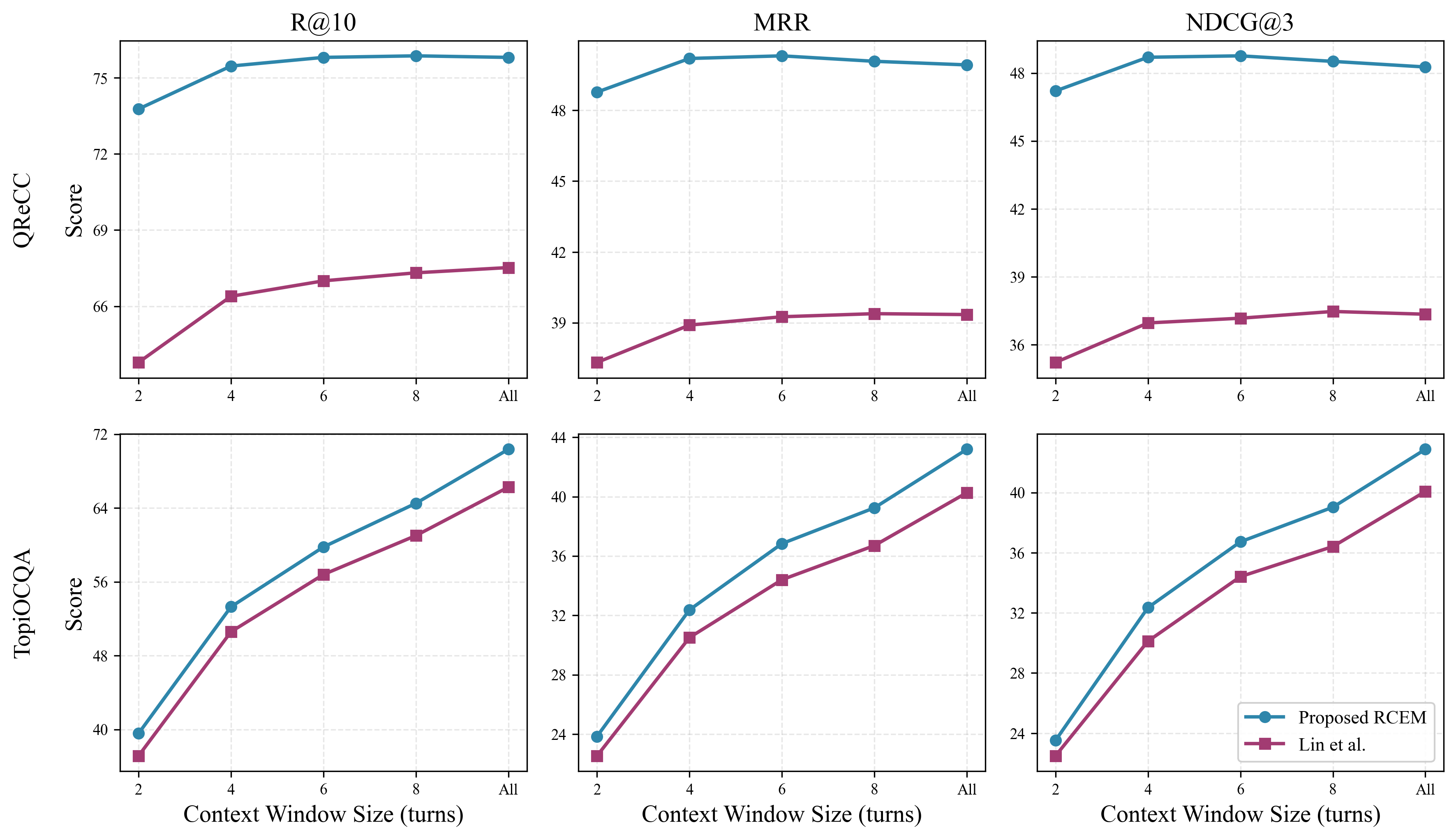}
\caption{{\bf Accuracy comparison VS context window size.} On the TopiOCQA benchmark, the retrieval accuracy of RCEM improves more with additional conversational turns, suggesting that RCEM is more effective at identifying and leveraging relevant information from long conversation histories.}
\label{fig:ctx}
\end{figure*}

\paragraph{Experiment setting.}
To evaluate RCEM, we use widely adopted conversational search benchmarks: QReCC~\cite{qrecc}, TopiOCQA~\cite{topiocqa}, and TREC CAsT~\cite{treccast2019, treccast2020}. QReCC and TopiOCQA provide both train and test splits, whereas TREC CAsT is primarily used for zero-shot evaluation. QReCC contains approximately 14K conversations and 80K question-answer pairs, with retrieval performed over a corpus of roughly 54M passages. TopiOCQA comprises 3.9K conversations and 51K question-answer pairs, paired with a corpus of approximately 25M passages. Together, these benchmarks enable evaluation of RCEM in both in-domain and distribution-shifted conversational retrieval settings. For consistency with prior work~\cite{adarewriter, emnlp2025, emnlp2021}, we report Recall@10, MRR, and NDCG@3 using \verb+pytrec_eval+~\cite{trec_eval}.

We compare RCEM against state-of-the-art CDR~\cite{emnlp2021, emnlp2025} and CQR~\cite{mo-etal-2023-convgqr, mao-etal-2023-llm4cs, mo-etal-2024-chiq, lai-etal-2025-adacqr, adarewriter} methods. Unless otherwise specified, both RCEM and~\citet{emnlp2021} use Qwen3-8B~\cite{qwen3} as the original embedder. RCEM uses GPT-4.1 to generate rewritten queries for training with a simple query-rewriting prompt (Appendix~\ref{app:prompt}). For a fair comparison, we also evaluate a Query Rewrite $\rightarrow$ Embedding pipeline using the same LLM (prompt) and embedding model as RCEM. All models are trained with a batch size of 32 on eight H100 GPUs. Unlike prior methods~\cite{emnlp2021, emnlp2025}, RCEM does not require conversation-to-passage relevance labels for training in any of the experiments. Additional implementation details are provided in Appendix~\ref{app:details}.

% We compare with the state-of-the-art CDR~\cite{emnlp2021, emnlp2025} and CQR~\cite{mo-etal-2023-convgqr, mao-etal-2023-llm4cs, mo-etal-2024-chiq, lai-etal-2025-adacqr, adarewriter} methods. Unless otherwise stated, RCEM and ~\citet{emnlp2021} use Qwen3-8B~\cite{qwen3} as original base embedder and RCEM uses GPT-4.1 as the LLM to rewrite query for training with simple query-rewriting prompt (Appendix~\ref{app:prompt}). Query rewriting with GPT-4.1 (with the same prompt to RCEM) $\rightarrow$ embedding (Qwen-8B) baseline is also compared for fair comparison.
% Models are trained with a batch size of 32 on 8 H100 GPUs. Unlike prior methods~\cite{emnlp2021, emnlp2025}, RCEM does not use mapping label between conversational queries and relevant passages in all the experiments. See more details of implementation in Appendix~\ref{app:details}.

% which are often difficult to obtain with high quality. Instead, it only requires LLM-generated rewritten queries from conversational inputs. This provides convenient and low-cost training labels, enabling RCEM to scale more easily to large-scale training.

\paragraph{In domain experiments.}
We conduct in-domain evaluations on the QReCC and TopiOCQA datasets. For a fair comparison with~\citet{emnlp2021}, we keep the experimental setup largely identical, including the hardware configuration, adaptation architecture, and number of training epochs. The primary difference lies in the training objective. We also compare against a Query Rewrite $\rightarrow$ Embedding baseline that uses the same GPT-4.1 query rewriter and Qwen3-8B embedder as RCEM. As shown in Table~\ref{tab:larger-model-indomain}, RCEM consistently outperforms prior methods, achieving $\sim 10\%$ and $\sim 13\%$ higher MRR than~\citet{emnlp2021} and~\citet{emnlp2025} on QReCC, respectively. 

Notably, RCEM also surpasses the Query Rewrite $\rightarrow$ Embedding baseline despite using the same LLM and embedding models, which hypothetically suggests that treating the rewritten query as an intermediate textual representation may introduce information loss. In contrast, RCEM directly aligns conversational queries with the embedding representation of rewritten queries, producing a more faithful and effective retrieval representation.

\paragraph{Distributional shift experiments.}
In the distribution-shift experiments,~\citet{emnlp2021} exhibits substantial performance degradation, whereas RCEM remains robust under distribution shift. Table~\ref{tab:outdomain_8b} reports retrieval performance when models are trained on one dataset and evaluated on another under an identical experimental setup. RCEM consistently outperforms both~\citet{emnlp2021} and the Query Rewrite $\rightarrow$ Embedding baseline across all dataset combinations and evaluation metrics. In particular, when trained on QReCC and evaluated on TREC CAsT 2019,~\citet{emnlp2021} performs substantially worse than RCEM, with accuracy gaps often ranging from $10\%$ to $30\%$. Similar trends are observed when the same experiments are conducted using a smaller base embedder, Qwen3-0.6B (Appendix~\ref{app:exp}).

This performance degradation is likely attributable to the contrastive objective used by~\citet{emnlp2021}, which directly aligns long conversational queries with long passages. Such a learning objective is inherently more complex and can cause the model to overfit to the training dataset. In contrast, RCEM maps conversational inputs to shorter rewritten queries, providing a simpler learning target while preserving the original embedding space. Attempting to preserving the embedding space also acts as an effective regularizer, improving robustness to distribution shift.

% In the distributional shift experiments,~\citet{emnlp2021} exhibits a pronounced overfitting issue, whereas RCEM maintains robust performance even under distributional shift. Table~\ref{tab:outdomain_8b} reports the retrieval accuracy when models are trained on one dataset and evaluated on another under the same experimental setup for a fair comparison. RCEM consistently outperforms~\citet{emnlp2021} and and Query Rewrite $\rightarrow$ Embedding baseline across all dataset combinations and evaluation metrics. In particular, when~\citet{emnlp2021} is trained on QReCC and evaluated on TREC CAsT, its performance is substantially lower than that of RCEM, with gaps often ranging from $10\%$ to $30\%$. Similar results have been observed when running the same experiments with smaller original embedding model, Qwen3-0.6B (See Appendix~\ref{app:exp}).

% This performance degradation is likely due to the contrastive objective used by~\citet{emnlp2021}, which directly maps long conversational queries to long passages. This task is more complex and can cause the model to overfit to the training dataset. In contrast, RCEM maps conversational inputs to shorter rewritten queries, which provides a simpler learning target while preserving the original embedding space. Preserving the embedding space serves as an effective regularizer, helping to improve robustness to distributional shift.

\paragraph{Accuracy VS context window.}
We further conduct an ablation study to examine the impact of conversational history length on retrieval performance. Figure~\ref{fig:ctx} shows that RCEM consistently outperforms~\citet{emnlp2021} across different context window sizes, with accuracy gains increasing as more conversational history is provided. In particular, RCEM benefits more from longer contexts on TopiOCQA, suggesting that it is better able to capture and leverage salient information distributed across multi-turn conversations.

% We also perform an ablation study to evaluate the effect of conversational history length on retrieval accuracy. Figure~\ref{fig:ctx} shows that RCEM consistently outperforms~\citet{emnlp2021} across different context window sizes and achieves larger gains as more conversational history is provided. In particular, on TopiOCQA, RCEM benefits more from longer contexts, indicating RCEM's ability to better capture and utilize salient information dispersed across multi-turn conversations.

\begin{table*}[t]
\centering
\small
\begin{tabular}{llcccc}
\toprule
Type & Method & R@10 & MRR & NDCG@3 & Latency\\
\midrule
CQR & AdaRewriter, $N=16$~\cite{adarewriter}
    & 69.8 & 47.5 & 44.7& 3,656 ms\\
CQR &Query Rewrite (GPT-4.1) with baseline prompt $\rightarrow$ Embedding 
    & 73.3 & 47.3 & 45.4& 1,275 ms\\
CQR &Query Rewrite (GPT-4.1) with~\citet{mao-etal-2023-search} $\rightarrow$ Embedding 
    & 75.1 & 49.3 & 47.5& 2,495 ms\\
CDR &\citet{emnlp2021}
    & 67.5 & 39.3 & 37.4& 69 ms\\
CDR &Proposed RCEM with baseline prompt
    & 75.8 & 49.9 & 48.3& 69 ms\\
CDR &Proposed RCEM with ~\citet{mao-etal-2023-search}
    & \textbf{78.1} & \textbf{52.3} & \textbf{50.8}& 69 ms\\
\bottomrule
\end{tabular}
\caption{\textbf{Effect of rewrite quality on RCEM and latency.} When RCEM is trained with a stronger query rewriting method~\cite{mao-etal-2023-search}, retrieval accuracy on QReCC dataset is better than baseline prompt setting, while maintaining the latency in inference.}
\label{tab:rewrite-quality}
\end{table*}

\begin{table}[t]
\centering
\small
\begin{tabular}{llccc}
\toprule
Benchmark & Loss & R@10 & MRR & NDCG@3 \\
\midrule
\multirow{2}{*}{QReCC}
 & SL  & {\bf 75.8} & {\bf 49.9} & {\bf 48.3} \\
 & $L_2$ & 74.4 & 47.2 & 45.2 \\
\midrule
\multirow{2}{*}{TopiOCQA}
 & SL & {\bf 70.4} & {\bf 43.2} & {\bf 42.9} \\
 & $L_2$ & 53.9 & 30.4 & 29.7 \\
\bottomrule
\end{tabular}
\caption{{\bf Loss comparison.} Structural Learning (SL) loss proposed in RCEM consistently better performs than $L_2$ loss used in~\citet{emnlp2025} on both QReCC and TopiOCQA datasets.}
\label{tab:loss}
\vspace{0.3em}
\end{table}

\paragraph{Structure learning vs $L_2$.}
% One of the differentiator of RCEM from ~\cite{emnlp2025} (use $L_2$ loss) is the Structure Learning loss that minimize point-wise distance and pairwise distance. The Table~\ref{tab:loss} consistently show that Structural Learning (SL) loss proposed in RCEM is consistently better perform than $L_2$ loss used in~\citet{emnlp2025} on QReCC and TopiOCQA datasets.

A key differentiator of RCEM from~\citet{emnlp2025} is the proposed Structure Learning (SL) loss, which preserves both pointwise embedding similarity and pairwise structure during training. Table~\ref{tab:loss} shows that the SL loss consistently achieves better retrieval performance than the $L_2$ loss used in~\citet{emnlp2025} on both QReCC and TopiOCQA.

\begin{figure}[t]
\centering
\includegraphics[width=0.9\columnwidth]{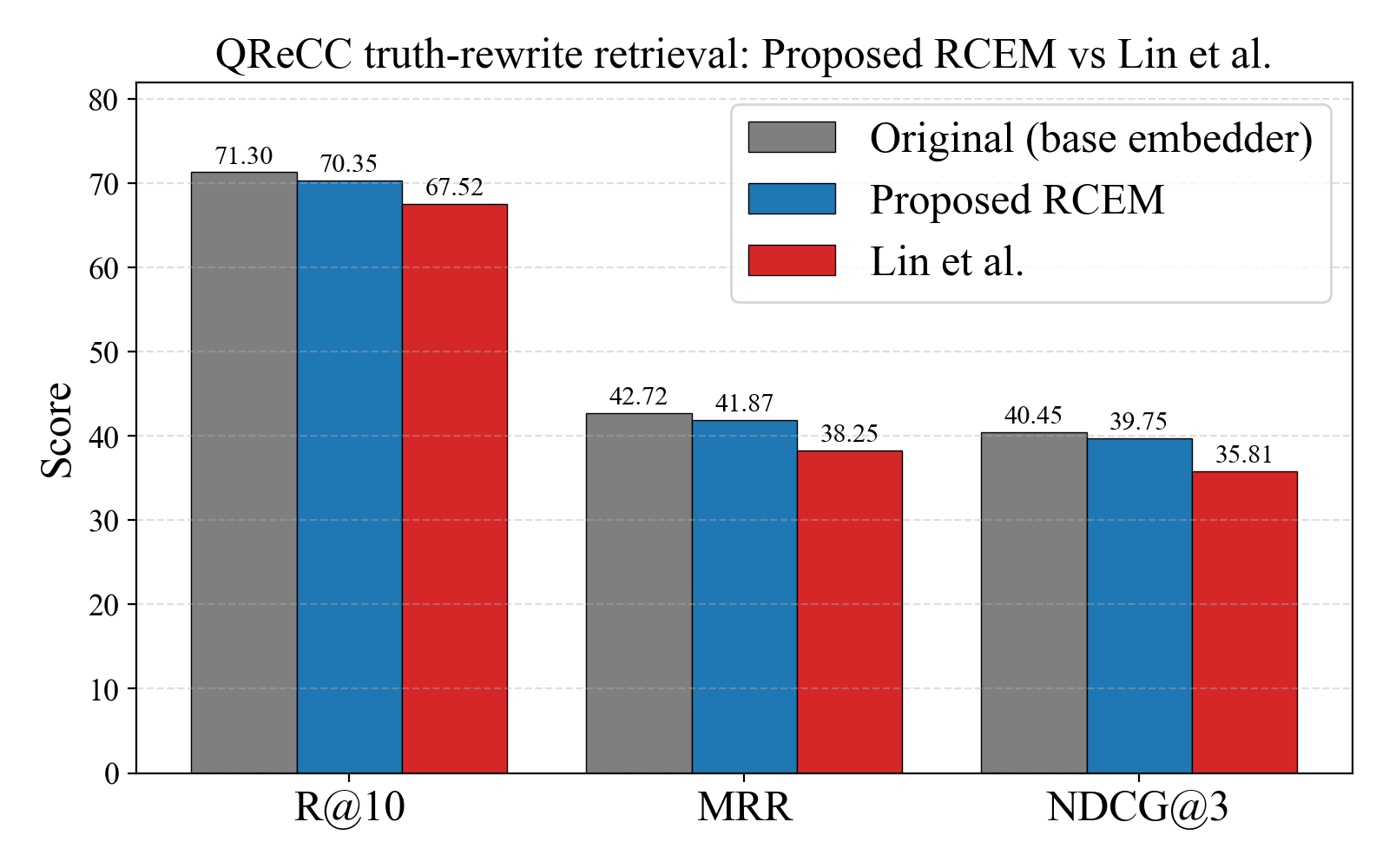}
\caption{{\bf Approximately preserving original embedding space.} General query is embedded by RCEM and run retrieval on index built from the original embedding. RCEM better preserves retrieval accuracy than ~\citet{emnlp2021} compared with the original embedder. The experiment is performed with Qwen3-0.6B base embedder.}
\label{fig:preserve}
\end{figure}

\paragraph{Approximately preserving original embedding space.}
Figure~\ref{fig:preserve} shows whether RCEM preserves the original embedding space. Specifically, we encode general queries (the GT rewrite-query in QReCC) using RCEM and perform dense retrieval over a QReCC index built by the original embedder. We then compare the retrieval accuracy of RCEM against the original embedder and~\citet{emnlp2021}. The results show that RCEM experiences a small performance drop compared to the original embedder, whereas~\citet{emnlp2021} suffers a larger degradation. We attribute this difference to the fact that~\citet{emnlp2021} does not explicitly preserve the original embedding space during training.

% We evaluate whether RCEM preserves the original embedding space. Specifically, we encode general queries (groundtruth standalone rewrite in QReCC) using RCEM and perform dense retrieval on an QReCC index built by the original embedder. We then compare retrieval accuracy against the original embedder and ~\cite{emnlp2021}. The results show that RCEM retrieval accuracy is dropped compared with the original embedder but the degeneration is less than ~\citet{emnlp2021} because ~\citet{emnlp2021} does not attemp to preserve in the embedding space during training.

\paragraph{Effect of better rewrite quality and latency.}
% We compare performance of RCEM and two-step baseline methods with simple baseline prompt vs better query rewrite method~\cite{mao-etal-2023-search} on QReCC in Table~\ref{tab:rewrite-quality}. The stronger query rewrite method improves both the two-step baseline and RCEM. confirming that rewrite quality affects the final model. Under the same stronger rewriting strategy, RCEM still exceeds the two-step pipeline. Note that RCEM uses the query rewriting only in training time and the inference cost and latency maintained, whereas the two-step baseline increase cost and latency in inference time.

Table~\ref{tab:rewrite-quality} compares RCEM with the Query Rewrite $\rightarrow$ Embedding baseline on QReCC using either a simple prompting strategy or a stronger query rewriting method~\cite{mao-etal-2023-search}. The improved rewriting method enhances the retrieval accuracy of both approaches, indicating that rewrite quality has a direct impact on retrieval effectiveness. Nevertheless, RCEM consistently outperforms the two-step pipeline under the same rewriting strategy. Notably, RCEM uses query rewriting only during training so that better prompting strategy adds up better retrieval accuracy with inference latency unchanged. In contrast, the two-step baseline requires query rewriting at inference time, meaning that improved rewrite quality often comes at the cost of higher computational overhead and increased latency.

Table~\ref{tab:rewrite-quality} also compares inference latency. In general, conversational dense retrieval (CDR) methods exhibit lower latency than conversational query reformulation (CQR) methods because CDR do not require LLM calls during inference. RCEM achieves latency comparable to existing CDR methods while delivering higher accuracy than both CDR and CQR baselines. Additional details on computational cost are provided in Appendix~\ref{app:computational-cost}.

% Table~\ref{tab:rewrite-quality} shows RCEM and Query Rewrite $\rightarrow$ baselines on QReCC using either a simple prompting strategy or a stronger query rewriting method~\cite{mao-etal-2023-search}. The improved rewriting method boosts the accuracy of both approaches, indicating that query rewrite quality has a direct impact on retrieval effectiveness. Nevertheless, RCEM consistently outperforms the two-step pipeline under the same rewriting strategy. Note that query rewriting is used only during RCEM training, so its inference latency remain unchanged. In contrast, the two-step baseline requires query rewriting at inference time, meaning that improvements in rewrite quality often come at the expense of higher computational cost and increased latency. Latency in inference time is also compared in the Table~\ref{tab:rewrite-quality}. Generally, CDR methods presents less latencies because CDR does not require to call LLM model in inference time but CQR does not. RCEM presents similar latency while presenting better retrieval accuracy compared both with CQR and CDR. See more details in Section~\ref{app:computational-cost}.
\section{Conclusion}
\label{sec:conclusion}
We present RCEM, a conversational search embedder that implicitly maps conversational queries to LLM-rewritten queries and generates their embeddings without requiring explicit LLM-based query rewriting during inference. Unlike prior methods that address the challenging task of directly aligning long conversational contexts with long passages, RCEM maps conversational queries to shorter LLM-rewritten queries while attempting to preserve the original embedding space, thereby improving generalization and remaining compatible with existing retrieval indexes built by the original embedder. In addition, RCEM does not require expensive conversation-to-passage relevance supervision. Extensive experiments demonstrate that RCEM achieves strong retrieval accuracy and improved robustness under distribution shift while substantially reducing inference latency.

\section*{Limitations}
The primary limitation of RCEM lies in the LLM’s query rewriting capability, as the model is designed to distill LLM-based rewriting behavior into the embedder. Although the results in Table~\ref{tab:larger-model-indomain} show that RCEM consistently outperforms the two-step pipeline (Query Rewrite $\rightarrow$ Embedding), hypothetically because it learns a richer latent representation than explicit rewritten text, its accuracy still depends on the quality of the rewritten queries. As shown in Table~\ref{tab:rewrite-quality}, stronger query rewriting methods lead to higher retrieval accuracy for RCEM. Therefore, further improvements in query rewriting, such as incorporating advanced rewriting techniques~\cite{adarewriter} or leveraging multiple query rewrites as additional conversation-to-rewrite training pairs, could further enhance RCEM's retrieval accuracy.

%The main bottleneck of the proposed method is the LLM’s query rewriting capability because RCEM goals to inject the LLM-rewriting capability. Although experiments (Table~\ref{tab:larger-model-indomain}) show that RCEM outperforms the two-step baseline (LLM-based query rewriting followed by embedding), hypothetically because it learns a stronger latent representation than the explicit rewritten text, RCEM's performance still depends on the quality of query rewriting as shown in the experiments Table~\ref{tab:rewrite-quality} that show RCEM shows better retrieval accuracy with better query rewriting method. Improving the LLM’s rewriting, by leveraging prior rewriting methods~\cite{adarewriter} or other multiple rewrites could be incorporated as additional conversation-rewrite training pairs, can further enhance the performance of RCEM.

% Entries for the entire Anthology, followed by custom entries
\bibliography{anthology,custom}
\bibliographystyle{acl_natbib}

\clearpage      % Flush all remaining floats and force a new page
\onecolumn      % Switch the layout to a single column
\appendix
\newpage
\section{Latency and Computational Cost Analysis}
\label{app:computational-cost}

We evaluate inference latency and estimate recurring serving cost in Table~\ref{tab:computational-cost}. Following~\citet{emnlp2025}, latency is measured over the complete QReCC test set including all conversations. NVIDIA H100 GPU is used for the evaluation. We use the following public serving prices:
\begin{itemize}
    \item GPT-4.1: \$2 per million input tokens and \$8 per million output
    tokens from Azure OpenAI Service pricing\footnote{\url{https://azure.microsoft.com/en-us/pricing/details/azure-openai/}}
    (accessed August 2, 2026).
    \item Llama-3.1-8B-Instruct: \$0.02 per million input tokens and \$0.04
    per million output tokens from OpenRouter (DeepInfra)\footnote{\url{https://openrouter.ai/meta-llama/llama-3.1-8b-instruct\#providers}}
    (accessed August 2, 2026).
    \item Qwen2.5-7B-Instruct: \$0.01 per million tokens from OpenRouter
    (Phala)\footnote{\url{https://openrouter.ai/qwen/qwen-2.5-7b-instruct}}
    (accessed August 2, 2026).
    \item Qwen3-Embedding-8B: \$0.01 per million input tokens from OpenRouter
    (DeepInfra)\footnote{\url{https://openrouter.ai/qwen/qwen3-embedding-8b\#providers}}
    (accessed August 2, 2026).
\end{itemize}

\begin{table*}[t]
\centering
\small
\begin{tabular}{llrrr}
\toprule
Type & Method & Latency & Cost/1M Qry.& MRR \\
\midrule
CQR & Query Rewrite (GPT4.1) $\rightarrow$ Embedding (Qwen3-8B)
    & 1,275 ms & \$570.4 & 47.3 \\
CQR & Query Rewrite (GPT4.1) + ~\citet{mao-etal-2023-search} $\rightarrow$ Embedding (Qwen3-8B)
    & 2,495 ms & \$840.4 & 49.3 \\
CQR & AdaRewriter, $N=16$~\cite{adarewriter} (Llama-3.1-8B)
    & 3,656 ms & \$20.8 & 47.5 \\
\midrule
CDR & \citet{emnlp2021} (Qwen3-8B)
    & 69 ms & \$1.77 & 39.3 \\
CDR & \citet{emnlp2025} (Qwen2.5-7B)
    & \textbf{41 ms} & \$1.77 & 36.8 \\
CDR & Proposed RCEM (Qwen3-8B)
    & 69 ms & \$1.77 & 49.9 \\
CDR & Proposed RCEM + ~\citet{mao-etal-2023-search} (Qwen3-8B)
    & 69 ms & \$1.77 & \textbf{52.3} \\
\bottomrule
\end{tabular}
\caption{\textbf{Inference efficiency and accuracy on QReCC.} Latency, estimated recurring
serving cost, and retrieval performance for conversational query
reformulation (CQR) and conversational dense retrieval (CDR) methods.}
\label{tab:computational-cost}
\end{table*}
RCEM achieves higher retrieval accuracy than existing CDR methods while maintaining comparable latency and serving cost. Compared with the Query Rewrite $\rightarrow$ Embedding pipeline, RCEM reduces latency from $1,275$ ms to $69$ ms and lowers the estimated serving cost from $\$570.4$ to $\$1.77$ per million queries, while improving MRR from $47.3$ to $49.9$. RCEM also outperforms ~\citet{adarewriter} in retrieval accuracy while requiring substantially lower latency and serving cost.

Stronger offline query rewriting methods~\cite{mao-etal-2023-search} further improve RCEM's MRR from $49.9$ to $52.3$ without increasing inference latency or serving cost. This is possible because the improved query rewriting is used only during offline training-data generation and is never invoked at inference time.

% RCEM achieves higher retrieval accuracy than the other CDR methods at
% comparable latency and serving cost. Compared with Query Rewrite $\rightarrow$ Embedding, RCEM reduces latency from $1,275$~ms to $69$~ms and estimated
% cost from $\$570.4$ to $\$1.77$ per million queries while improving MRR from $47.3$
% to $49.9$. RCEM also outperforms AdaRewriter~\citet{adarewriter} in retrieval accuracy while
% requiring substantially less latency and serving cost.

% Stronger offline query rewriting method~\cite{mao-etal-2023-search} further improves
% RCEM's MRR from $49.9$ to $52.3$ without changing online inference latency or
% serving cost. This is possible because the stronger offline query rewriting~\cite{mao-etal-2023-search} is used only during offline training-data generation and is not invoked at inference time.

\newpage

\section{Baseline Prompt}
\label{app:prompt}
We use a simple prompt to convert each conversational query into a rewritten query for generating the training dataset, as shown below:
% \newsavebox\myv
% \begin{lrbox}{\myv}\begin{minipage}{\textwidth}
\begin{verbatim}
    Given the recent conversation:
    {conversation_history}
    
    Rewrite the following ambiguous query
    into a self-contained and more clear query.
    
    This is the ambiguous query:
    {query}
    
\end{verbatim}
% \end{minipage}\end{lrbox}
% \resizebox{0.9\textwidth}{!}{\usebox\myv}
% \\ \\

We enforce GPT4.1 to follow schema to reject redundant output token such as \begin{verbatim}
    The rewritten query is ...
    The clear query is ...
    
\end{verbatim}

% RCEM could potentially achieve even better performance by incorporating improved prompts or advanced query rewriting methods from prior work~\cite{adarewriter}.
\newpage

\section{More Experiments}
\begin{figure}[t]
\centering
\includegraphics[width=0.9\columnwidth]{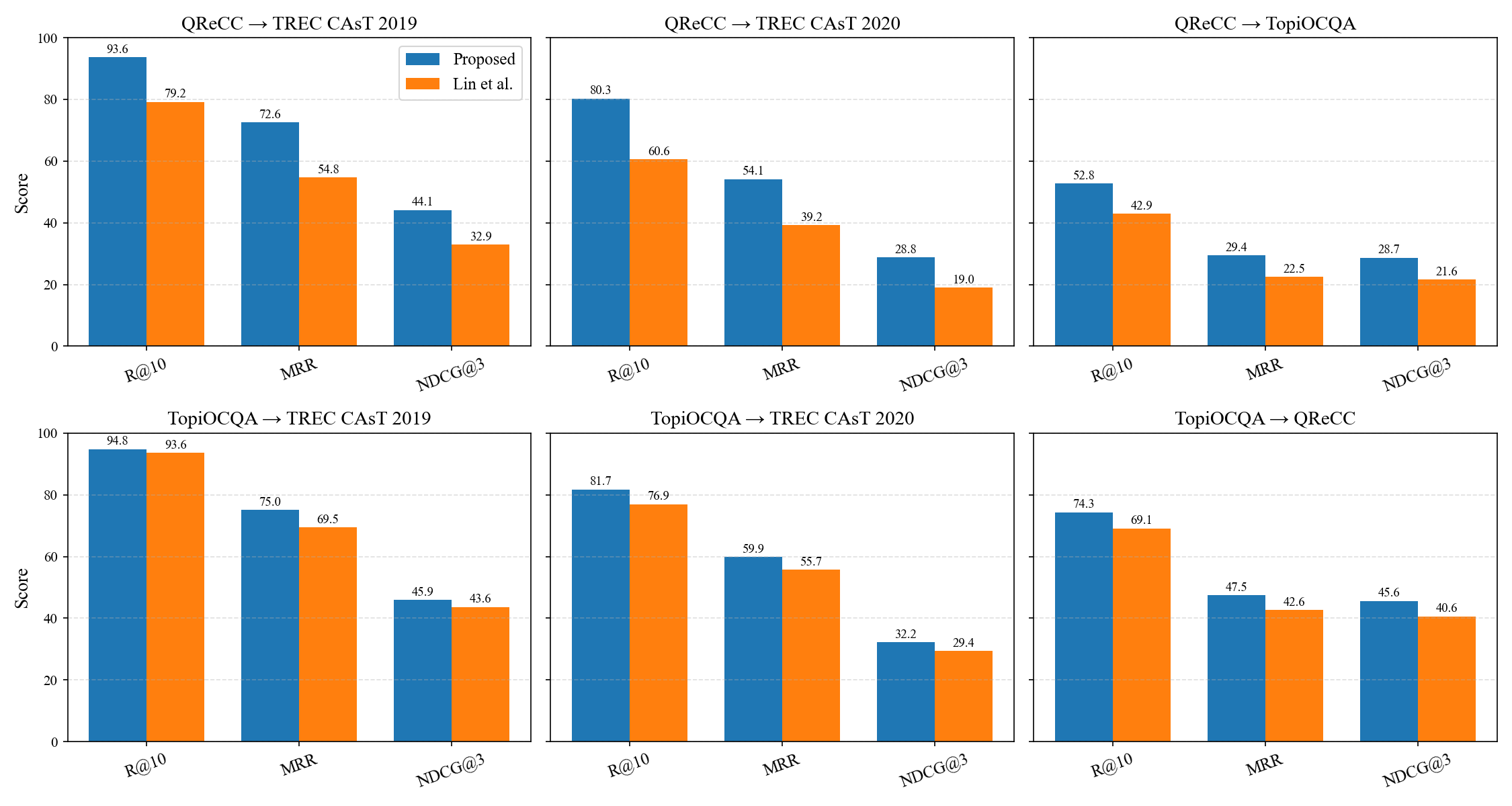}
\caption{{\bf Retrieval accuracy comparison in distributional shift with 0.6B Qwen3 base embedder}: We train the models on one dataset and evaluate them on another under the same experimental setup to ensure a fair comparison. The proposed RCEM method maintains strong performance under distributional shift, whereas the performance of~\citet{emnlp2021} drops noticeably. In particular, when trained on QReCC and tested on TREC CAsT 2020, RCEM achieves approximately $20\%$ higher Recall@10 than~\citet{emnlp2021}. Across all dataset combinations and evaluation metrics, RCEM consistently outperforms~\citet{emnlp2021}.}
\label{fig:outdomain_06b}
\end{figure}

We perform the distributional shift experiments with 0.6B Qwen3 base model. The proposed RCEM method maintains strong performance under distributional shift, whereas the performance of~\citet{emnlp2021} drops noticeably wtih 0.6B Qwen3 base model experiments. See Figure~\ref{fig:outdomain_06b}.
\label{app:exp}

\begin{table}[t]
\centering
\small
\begin{tabular}{rccc}
\toprule
$\lambda$ & R@10 & MRR & NDCG@3 \\
\midrule
0    & 53.9 & 30.4 & 29.7 \\
0.25 & 69.8 & 42.2 & 41.8 \\
0.5  & 69.9 & 42.4 & 42.2 \\
\textbf{1.0} & \textbf{70.4} & \textbf{43.2} & \textbf{42.9} \\
2.0  & 68.8 & 42.1 & 41.7 \\
4.0  & 57.9 & 33.5 & 33.2 \\
\bottomrule
\end{tabular}
\caption{\textbf{Sensitivity to the structure-learning weight.} TopiOCQA
retrieval performance as the pairwise structure-preservation weight
$\lambda$ varies. The setting used in the main experiments is shown in bold.}
\label{tab:structure-learning-sensitivity}
\end{table}

We analyze the sensitivity of RCEM to $\lambda$, which controls the weight of
the pairwise structure preservation objective relative to the pointwise
alignment objective. Table~\ref{tab:structure-learning-sensitivity} reports
retrieval performance on TopiOCQA. Performance remains stable over the broad range $\lambda \in [0.25, 2.0]$:
R@10 varies by only $1.5$ points, MRR by $1.1$ points, and NDCG@3 by $1.2$
points. The value used in the main experiments, $\lambda=1.0$, achieves the
best result on all three metrics.

The extreme settings further clarify the role of pairwise structure
preservation. Setting $\lambda=0$ removes the pairwise objective and reduces
R@10, MRR, and NDCG@3 by $16.5$, $12.8$, and $13.2$ points, respectively,
relative to $\lambda=1.0$. Conversely, $\lambda=4.0$ overemphasizes pairwise
geometry at the expense of pointwise conversational-to-rewrite alignment.
These results indicate that pairwise structure preservation is important,
while RCEM is not sensitive to moderate choices around the selected value.

\clearpage
\newpage

\section{More Comparison with Prior Works}
\label{app:compare}
\begin{figure}[t!]
\centering
\includegraphics[trim=3cm 7cm 4cm 0, width=0.8\columnwidth]{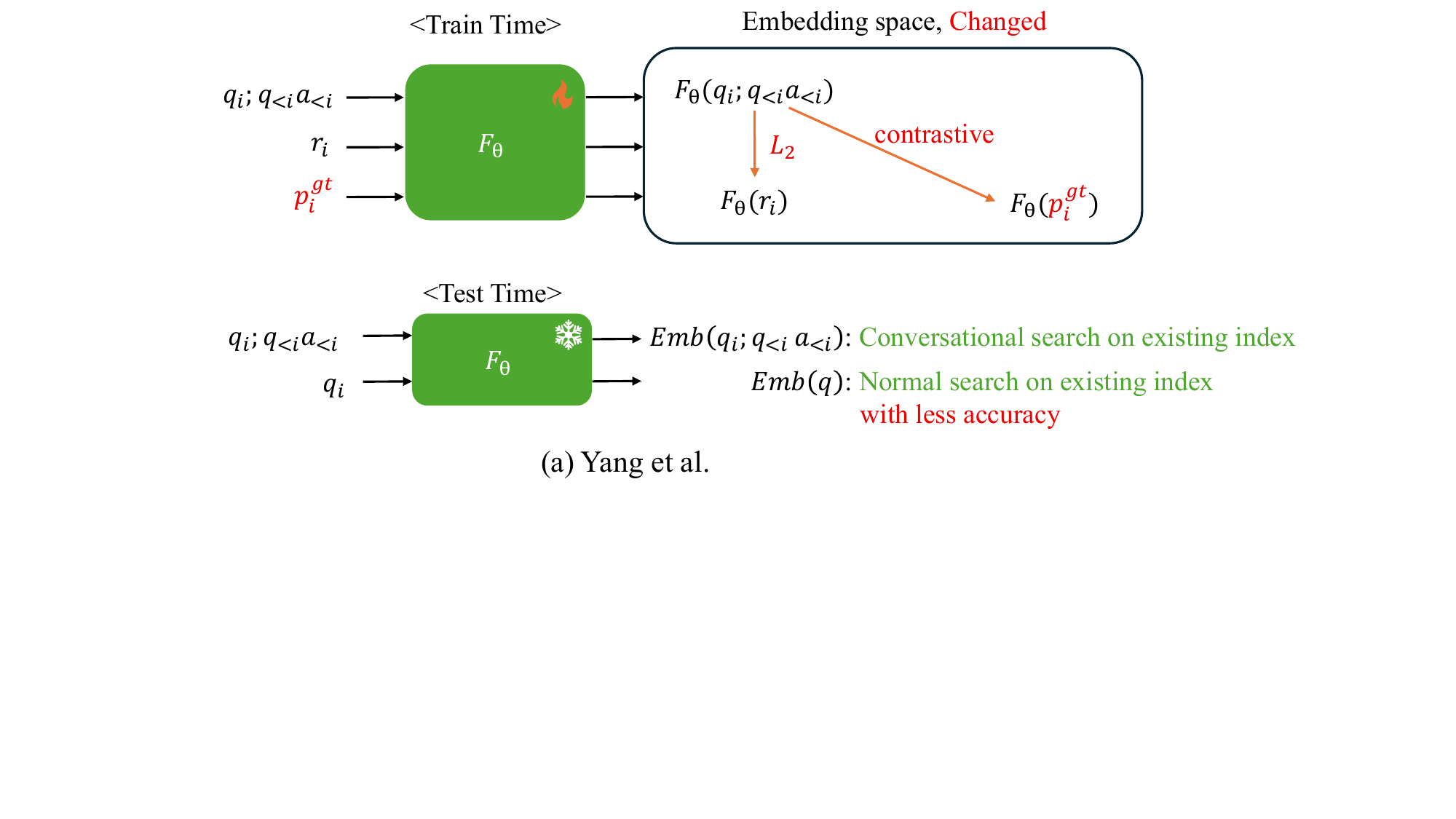}
\includegraphics[trim=3cm 3cm 4cm 0, width=0.8\columnwidth]{figs/compare2.pdf}
\caption{{\bf RCEM, comparison with prior work.} 
% RCEM first precomputes embeddings using the original frozon embedder $G$. Specifically, it embeds the LLM-rewritten query, $r_i = LLM(q_i; q_{<i}a_{<i})$, as well as general queries $q_i$ and passages $p_i$. 
%, generated from the current query and conversation history using LLM 
% The RCEM $F_\theta$, is then trained to map the conversational query with a special token $F_\theta(c_i)$, to the embedding of the rewritten query embedded by the frozen embedder $G(r_i)$ using the proposed Structure Learning (SL) loss. During training, RCEM also aligns general queries and documents with their precomputed embeddings, i.e., $F_\theta(q_i) \rightarrow G(q_i)$ and $F_\theta(r_i) \rightarrow G(r_i)$, to preserve the original embedding space. This design enables a single model to support both conversational search, triggered by the special token, and standard search over the existing index. Preserving embedding space also serves as a regularizer, making the model more robust to distributional shift. 
Different from RCEM, ~\citet{emnlp2025} trains the model to map the conversational query embedding $F_\theta(q_i; q_{<i}a_{<i})$ directly to the ground-truth passage $G(p_i^{gt})$ with contrastive loss and also maps to rewritten query $F_\theta(r_i)$ with $L_2$ loss.~\citet{emnlp2025} requires conversational queries and relevant passages for their training. Because~\citet{emnlp2025} does not use embeddings produced by the original encoder $G$ as training anchors, the learned representations can drift from the original embedding space. This embedding-space shift may degrade compatibility with existing document indexes encoded by $G$, whereas RCEM is designed to preserve embedding-space compatibility while adapting conversational queries.
% RCEM does not require conversational-query-to-ground-truth-passage mappings, which are often difficult to obtain accurately. This makes RCEM more scalable, as it reduces the need for expensive and high-quality relevance annotations
}
\label{fig:compare_yang}
\end{figure}

We compare RCEM with ~\citet{emnlp2025} in this Section. See Figure~\ref{fig:compare_yang} for more details.

\newpage
\clearpage
\section{More Implementation Details}
\label{app:details}
RCEM is trained using the Unsloth training framework~\cite{unsloth2024}. The head projects the base embedding into an intermediate representation of size $3\times1024$, followed by a projection back to the 1024-dimensional embedding space. Note that we only use the first 1024 embeddings from the base model. We use LoRA rank=$32$ and apply adapters to all attention and feed-forward projection layers, including \texttt{q proj}, 
\texttt{k proj}, \texttt{v proj}, \texttt{o proj}, \texttt{gate proj}, \texttt{up proj}, and \texttt{down proj}. Training is performed with the Adam optimizer using a learning rate of $3\times10^{-4}$, no weight decay, and a cosine learning-rate schedule with a warmup ratio of $0.1$. To preserve compatibility with the original retrieval embedding space, we jointly optimize the embedding preservation objective using a randomly sampled 1\% subset of passages, standalone queries, and conversational queries from the training data.

\end{document}